\documentclass[conference]{IEEEtran}
\IEEEoverridecommandlockouts
\usepackage{cite}
\usepackage{amsmath,amssymb,amsfonts}
\usepackage{algorithmic}
\usepackage{graphicx}
\usepackage{textcomp}
\usepackage{xcolor}
\usepackage{hyperref}
\usepackage{multirow}
\usepackage{balance}
\usepackage[moderate, tracking=normal]{savetrees}
\def\BibTeX{{\rm B\kern-.05em{\sc i\kern-.025em b}\kern-.08em
    T\kern-.1667em\lower.7ex\hbox{E}\kern-.125emX}}
    
\begin{document}

\title{Implicit Communication of Contextual Information in Human-Robot Collaboration}

\author{
\IEEEauthorblockN{ 
Yan Zhang}
\IEEEauthorblockA{
\textit{University of Melbourne}\\
Melbourne, Australia \\
yan.zhang.1@unimelb.edu.au
}
}

\maketitle

\begin{abstract}
Implicit communication is crucial in human-robot collaboration (HRC), where contextual information, such as intentions, is conveyed as implicatures, forming a natural part of human interaction. However, enabling robots to appropriately use implicit communication in cooperative tasks remains challenging. My research addresses this through three phases: first, exploring the impact of linguistic implicatures on collaborative tasks; second, examining how robots' implicit cues for backchanneling and proactive communication affect team performance and perception, and how they should adapt to human teammates; and finally, designing and evaluating a multi-LLM robotics system that learns from human implicit communication. This research aims to enhance the natural communication abilities of robots and facilitate their integration into daily collaborative activities.
\end{abstract}


\section{Introduction}
In human-robot interaction, the classification of explicit and implicit communication is often based on straightforward criteria. For example, language communication, which can be heard, is considered explicit interaction~\cite{bolt1980put}, while eye gaze, which is inconspicuous, is considered implicit interaction~\cite{li2017implicit}. However, I would like to differentiate them at a more profound and subtextual level, where implicit communication serves as a way to convey contextual information that receivers can interpret under mutual understanding. This view aligns with the broader nature of human communication, which is inherently versatile and relies heavily on subtle cues to convey intentions and achieve goals~\cite{frank2012predicting}. Known as ``implicatures'' in linguistics~\cite{grice1991studies}, this concept extends to various modalities beyond language; for example, applause can signal approval or subtly imply “Let’s end this”~\cite{gilbert2001joyful}. Thus, my dissertation defines implicit communication by the contextual information it conveys.

Implicit communication is an optimised way to communicate that commonly occurs in collaboration settings where teammates build a shared understanding of the task~\cite{clark1986referring}. A notable example is the use of indirect speech act, a pragmatic feature that requires the listener to infer the speaker's intended meaning from related context~\cite{searle1975indirect}. For instance, the statement ``I'm cold'' can imply a request to close the window, depending on the context. Previous work has shown that humans tend to use implicit ways to convey intentions when interacting with robots, at frequencies similar to those used with other humans~\cite{lee2010receptionist, bennett2017differences}. However, in real-world collaboration scenarios, the appropriateness of implicit intentions is less obvious due to the lack of clarity. There remains a gap in empirical evidence regarding understanding the nuanced role of implicit communication in HRC~\cite{brawer2018situated}. Therefore, the first research question I seek to address is: \textbf{(RQ1) How does the robot's ability to interpret implicature in language affect the perceived team performance and user experience in physical collaboration scenarios?}

In collaborative tasks, information exchange between teammates often occurs in a bidirectional flow~\cite{unhelkar2020decision, tian2023crafting}. While attention has been given to implicit interaction from humans to robots~\cite{mahadevan2021grip, knepper2017implicit, che2020efficient}, there is still a need for further exploration of how robots can convey implicit information to humans and how this affects HRC~\cite{fiore2013toward, lemasurier2021methods}. The potential for robots to communicate contextually relevant information implicitly, without interrupting the flow of interaction, presents a potential opportunity for enhancing the efficiency of HRC. This form of communication could also act as a backchannel, subtly responding to human's implicit intentions, helping to reduce ambiguity while maintaining fluent and natural interaction. Additionally, it is worth investigating whether robots should adapt their communication style based on how humans communicate—for instance, whether a robot should mirror implicit communication when the human initiates it. Further investigation into these domains could reveal strategies to enhance HRC, making it more natural and efficient. Therefore, the research questions in the second study I seek to address are: \textbf{(RQ2.1) What is the influence of robots using implicatures when conveying contextual information to humans during physical collaborative tasks?}, \textbf{(RQ2.2) How do different methods of implicit backchannel cues from robots influence human perception and team performance?}, and \textbf{(RQ2.3) Should robots adapt their communication style based on the human’s use of implicit/explicit communication?}

Building upon the understanding of implicit communication in human-robot collaboration, my research seeks to advance the integration of multimodal implicit cue interpretation and generation into robotic systems. By leveraging multimodal Large Language Models (LLM) that infer different modalities, such as speech, gaze, and motion, robots could potentially improve their ability to understand and respond to implicit intentions in a more nuanced manner. Additionally, by adapting to human behaviours and learning to generate appropriate implicit cues themselves, robots may become more adept at facilitating natural, bidirectional communication. However, relying on a single LLM presents several challenges. Due to knowledge gaps or misinformation, LLM outputs may suffer from hallucination or bias~\cite{ji2023survey, feng2023pretraining}. A cooperative approach that combines decisions from multiple LLMs could enhance the robustness and reliability of the system~\cite{feng2024don}. Furthermore, when a single LLM is tasked with complex scenarios, such as interpreting human language and gaze, and planning and executing assembly actions, extensive fine-tuning is required for different task components (e.g., perception, planning, execution), along with integration of external tools (e.g., API calls)~\cite{qin2023toolllm, zeng2023agenttuning}. By adopting a multi-LLM approach, where each LLM focuses on a specific component, robots could extend their implicit communication capabilities beyond basic tasks like object grasping to more complex, dynamic, and collaborative manipulations~\cite{kang2024prograsp} and also facilitate easier model updates for long-term use~\cite{shen2024small}. This approach could pave the way for a more seamless, intuitive HRC. Therefore, the third research question I seek to address is: \textbf{(RQ3) How can multi-LLMs be leveraged to enable robots to learn and enhance implicit communication abilities from humans in physical collaborative tasks?}

In summary, the contribution of this research lies in advancing the understanding and integration of implicit communication of contextual information in human-robot collaboration. I plan to conduct three major user studies to explore how robots' interpretation of human implicatures influences teamwork, how robots can effectively use implicatures to convey contextual information, and how the multi-LLM robotic system can enhance bidirectional communication in collaborative tasks. This research aims to improve HRC by reducing the cost and time required for ad-hoc human-robot team training, improving the naturalness of the interaction, and simplifying the building of common ground for better team efficiency. Advancing collaborative robots in physical tasks makes complex robots, like home service devices, easier to use, enabling effective deployment in healthcare, manufacturing, and home settings.

\section{Research Approach and Prior Work}
My research is based on lab empirical studies that involved a robot and a participant. I use TIAGo, a mobile manipulator robot with anthropomorphic features, including a 2 Degrees of Freedom head, neck, torso, and 7 Degrees of Freedom arm. TIAGo is equipped with an RGB-D camera, stereo microphone, and speaker that is suitable for human-robot interaction research and tabletop manipulation tasks~\cite{pages2016tiago}. I built a motion teleoperation interface and speech control software using the Robotics Operating System (ROS) and the TIAGo API. These customised tools can facilitate experiments to control the robot in pilots and user studies.

\begin{figure}
    \centering
    \includegraphics[width=1\columnwidth]{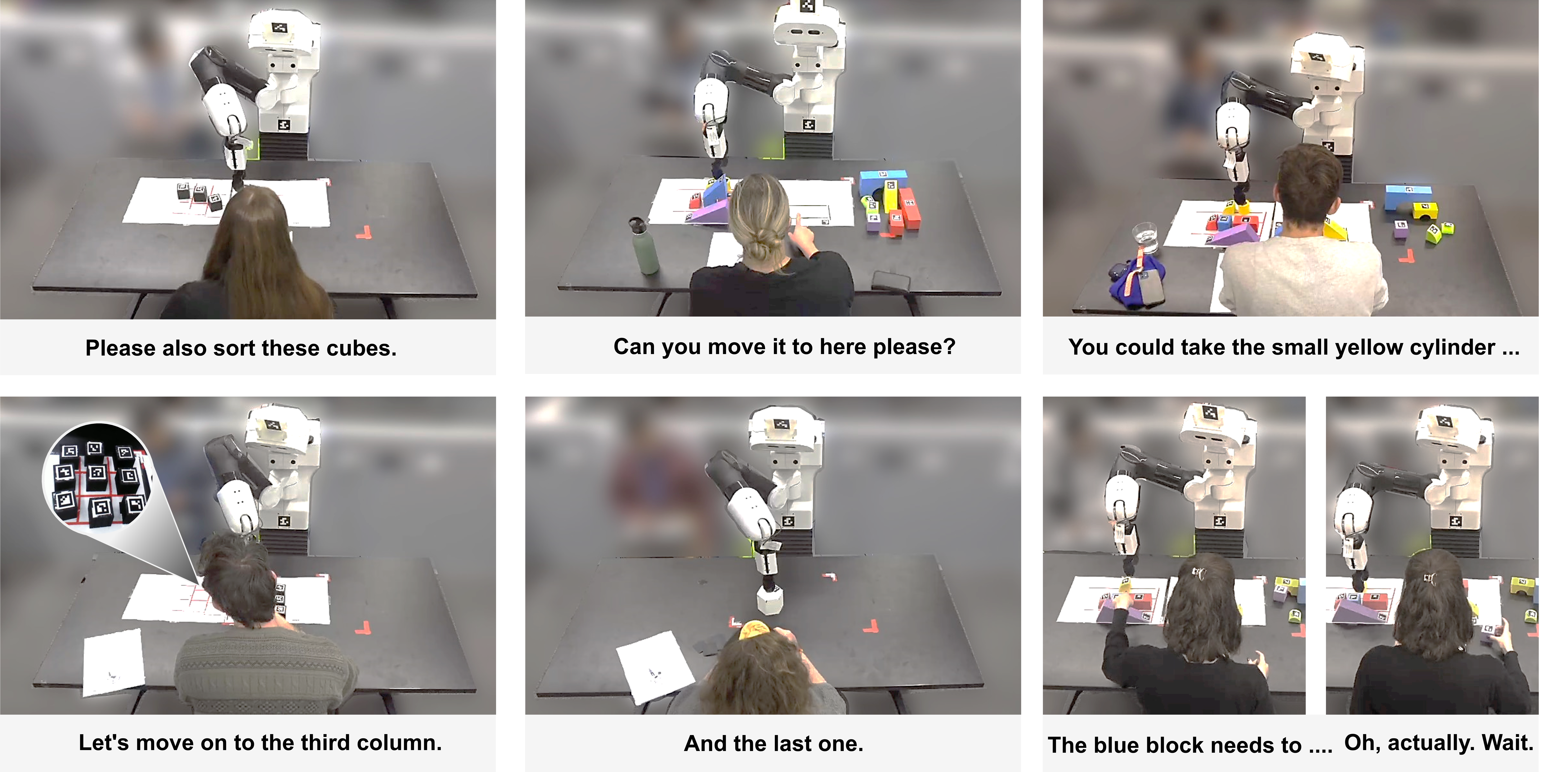}
    \caption{This figure shows images from Study 1, featuring representative participant utterances to illustrate the types of requests with implicit intentions. (Top left: explicit request; Others: implicit requests.)}
    \label{fig:study1}
\end{figure}

To answer RQ1, I conducted a lab study, engaging a participant and a robot in three collaborative physical tasks, to investigate the impacts of implicit language on human-robot collaboration. \autoref{fig:study1} shows photos from the lab study. The experiment involved 36 participants split into two groups: one interacted with a robot that understood implicatures, and the other with a robot responding only to direct commands. Participants completed three collaborative tasks: assembly, sorting, and polishing. Results showed that a robot's ability to interpret implicatures improved perceived team performance, trust, and anthropomorphism, fostering deeper cognitive engagement and a stronger sense of partnership. However, qualitative data indicated that implicature use can be task- and context-dependent, with inappropriate use potentially leading to negative impacts on trust. These insights emphasise the importance of using implicit communication in a contextually adaptive and appropriate manner. Therefore, the careful integration of explicit information and implicatures in verbal communication emerges as a critical factor in optimising the performance and overall experience of HRC.

\section{Future Work}
Based on the findings from Study 1, I am currently conducting a second user study to investigate the effects of implicit communication by robots on human perception and team performance during physical collaborative tasks. The study focuses on three key scenarios: 1) when the robot proactively uses implicatures to convey contextual information, 2) when the robot uses implicit cues as backchanneling to respond to human implied intentions, and 3) when the robot adapts its communication style in response to the human’s use of implicit or explicit communication. Specifically, I will compare four methods of communication: \textit{explicit speech}, \textit{implicit speech}, \textit{explicit motion}, and \textit{implicit motion}, across three scenarios: \textit{Human-lead (Robot-backchannel)}, \textit{Robot-lead (Robot-proactive)}, and \textit{Robot-adaptive (Robot mirrors human’s communication style)}. The study will measure team goal alignment, teamwork efficiency, fluency, and trust. Overall, the second study aims to explore the impact and potential of robots using implicit communication in collaborative tasks, providing valuable insights for designing robotic systems and interaction methods.

Building on the insights gained from Studies 1 and 2, the final phase of my research focuses on designing and evaluating a robotic system that employs cooperative multimodal multi-LLMs for human-robot collaborative manipulation tasks using the dataset collected in Study 1. In this system, each LLM will be initialised from a pre-trained base model and progressively fine-tuned for distinct roles, such as interaction perceiver, interaction generator, environment perceiver, task planner, and executor. The LLMs will cooperate to generate responses for each role, mitigating issues such as hallucinations. A user experiment will compare the performance of the multi-LLM system with that of a single LLM system on a simple pick-and-place task and a complex manipulation task. The system evaluation will include both quantitative metrics, such as response time and task success rate, and qualitative analysis of failure cases. Additionally, I will assess the interaction through quantitative measures like trust, team efficiency, and task fluency, alongside qualitative insights into human perception of the robot.

\section*{Acknowledgment}
Thank you to A/Prof.Wafa Johal, A/Prof.Jorge Goncalves, and Dr.Jarrod Knibbe for their guidance. I wish to acknowledge the contribution of the Melbourne Research Scholarship funded by the University of Melbourne.

\bibliographystyle{ieeetr}
\balance
\bibliography{ref}

\end{document}